\documentclass[conference]{IEEEtran}
\IEEEoverridecommandlockouts
\usepackage[font=small,skip=3pt]{caption}
\usepackage{cite}
\usepackage{amsmath,amssymb,amsfonts}
\usepackage{algorithmic}
\usepackage{graphicx}
\usepackage{textcomp}
\usepackage{xcolor}
\usepackage[hidelinks]{hyperref}
\def\BibTeX{{\rm B\kern-.05em{\sc i\kern-.025em b}\kern-.08em
    T\kern-.1667em\lower.7ex\hbox{E}\kern-.125emX}}
\begin{document}

\title{Improving Bidding and Playing Strategies in the Trick-Taking game Wizard using Deep Q-Networks}

\author{
\IEEEauthorblockN{Jonas Schumacher}
\IEEEauthorblockN{\textit{Nexocraft GmbH} \\
Bonn, Germany \\
jonas.schumacher@tu-dortmund.de}
\and
\IEEEauthorblockN{Marco Pleines}
\IEEEauthorblockA{\textit{Dpt. of Computer Science, TU Dortmund University} \\
Dortmund, Germany \\
marco.pleines@tu-dortmund.de}
}
\maketitle
\begin{abstract}
In this work, the trick-taking game \emph{Wizard} with a separate bidding and playing phase is modeled by two interleaved partially observable Markov decision processes (POMDP).
Deep Q-Networks (DQN) are used to empower self-improving agents, which are capable of tackling the challenges of a highly non-stationary environment. 
To compare algorithms between each other, the accuracy between bid and trick count is monitored, which strongly correlates with the actual rewards and provides a well-defined upper and lower performance bound.
The trained DQN agents achieve accuracies between 66\% and 87\% in self-play, leaving behind both a random baseline and a rule-based heuristic.
The conducted analysis also reveals a strong information asymmetry concerning player positions during bidding.
To overcome the missing Markov property of imperfect-information games, a long short-term memory (LSTM) network is implemented to integrate historic information into the decision-making process.
Additionally, a forward-directed tree search is conducted by sampling a state of the environment and thereby turning the game into a perfect information setting.
To our surprise, both approaches do not surpass the performance of the basic DQN agent.
\end{abstract}

\section{Introduction}

Deep Reinforcement Learning (DRL) has been shown to be a fruitful approach to optimize strategies in games with perfect information such as Atari games \cite{mnih2015human} or board games like Chess and Go \cite{silver2017mastering}.
At the same time, games with imperfect information like card games still constitute a major challenge for DRL algorithms \cite{sutton2018reinforcement}.
The partial observability violates the Markov property and is expected to require the integration of historic information to be solvable by DRL approaches \cite{morales2020grokking}.
Learning in a self-play multi-agent setting is highly non-stationary, which may cause troublesome instability during optimization.
Another challenge is apparent when the agent cannot take an action during the turn of another player, which does not conform to a regular Markov Decision Process (MDP).
Therefore, the agent has to operate on two different time scales.
One requires the agent to just perceive the other players' turns, while the other one asks the agent to take action.
Moreover, trick-taking games pose the issue of observation and action spaces that vary in size throughout a game.
At last, forward-directed search, common in two-person zero-sum games, cannot directly be applied in imperfect-information settings.

\IEEEpubid{\begin{minipage}{\textwidth}\ \\ \\[12pt] Accepted to IEEE CoG 2022 \end{minipage}}

\begin{figure*}
    \centering
    \includegraphics[width=.8\textwidth]{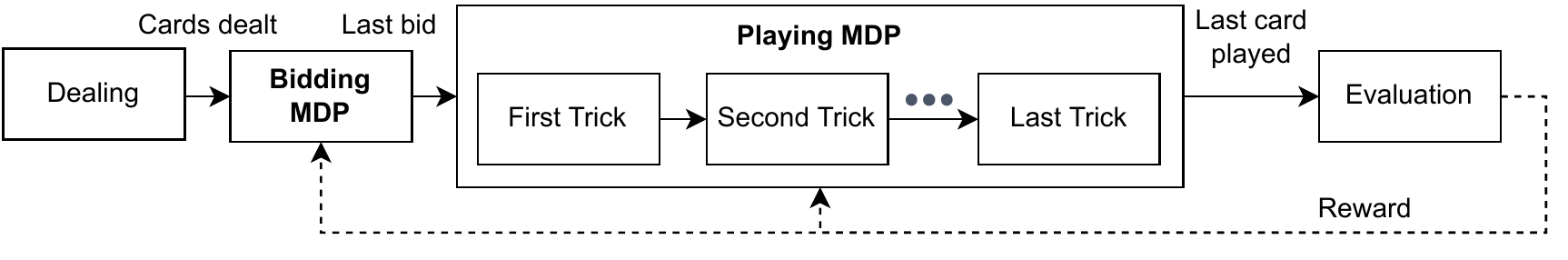}
    \caption{Structure of one game round: each round consists of a dealing, bidding, playing, and evaluation phase. The bidding and the playing phases are modeled by two interleaved (partially observable) MDPs.}
    \label{fig:structure_game_round}
\vspace{-0.15in}
\end{figure*}

This work studies the trick-taking game \emph{Wizard} with $4$ players, which consists of $15$ rounds shown in figure \ref{fig:structure_game_round}. Each round from $r=1$ to $r=15$ starts with a \emph{dealing} phase in which one card is drawn to determine the trump suit (except for the last round which has no trump suit) and exactly $r$ cards are dealt to each player. In the \emph{bidding} phase, the players need to estimate how many tricks they are going to take in the subsequent \emph{playing} phase. The first card in each trick sets the lead suit which has to be followed by the other players. The player who played the highest card wins the trick and begins the next one. Wizard is played with 60 cards consisting of a standard 52-card deck and 8 special cards. The 4 \emph{wizards} even beat the highest trump card and the 4 \emph{jesters} are beaten by even the lowest regular card. In the case of drawing one of the special cards in the beginning, it is assumed that no trump suit exists in that round. 
Only if players exactly hit their initial bid $x$, they receive $20 + 10\cdot x$ points in the \emph{evaluation} phase at the end of each round. If their final trick count $y$ differs from their initial bid $x$, they receive a penalty of $-10\cdot |x-y|$.


Due to the separate bidding and playing phase, Wizard is a highly interesting environment for the application of DRL algorithms. In bidding, players need to solve a prediction task by anticipating what is going to happen in the playing phase even though little information is available. The goal of this work is to not only create powerful self-improving agents but at the same time gain insights into the underlying imperfect-information environment itself.

We demonstrate that the Deep Q-Network (DQN) algorithm is able to handle the non-stationary environment induced by self-play.
The agent is robust and able to generalize by outperforming a random and rule-based agent, while only facing itself during training.
For the DQN implementation, neither supervised learning approaches nor human demonstrations are considered to establish this outcome.
To our knowledge, this work is the first one to use a purely DRL-based approach to optimize both bidding and playing.

We further study the integration of historic information to restore the violated Markov property.
A sequence-to-sequence model is trained in isolation to encode the entire history of played cards which is used to predict a diverse range of different tasks.
This model's hidden state is then used to augment the DQN agent's observation space, which surprisingly does not improve its performance.

At last, a model-based approach is examined. To this end, the agent learns to predict a probability distribution of possible true states of the environment.
It then samples a state from that distribution and performs a simple tree search to explore terminal nodes of the game tree to improve its decisions.
However, performing such a tree search does not increase the agent's outcome either.


This paper proceeds as follows.
Related work on similar trick-taking games and their approaches are showcased next.
In the approach section, the Wizard environment will be modeled by two interconnected POMDPs and the three approaches that seek to improve bidding and playing strategies will be introduced.
Afterward, results for all three approaches will be described.
The surprising ineffective outcomes of the two latter approaches and an apparent information asymmetry between players and across different game rounds are discussed before concluding this paper with future work.

\section{Related Work}

In the past years, several approaches were proposed to transfer the findings from perfect-information games to imperfect-information games with some of them focusing on trick-taking games in particular.
Yeh et al. \cite{yeh2018automatic} focused exclusively on the bidding phase of the cooperative card game \emph{Bridge}.
They implemented a DQN algorithm using raw data as input for the neural network.
Baykal et al. used fully-connected neural networks (FNN) to analyze the card game \emph{Batak}.
Supervised learning was used for bidding, and a Monte-Carlo-based Reinforcement Learning (RL) algorithm for playing \cite{baykal2019reinforcement}.
Another common approach in literature consists of imitating human expert moves by training the agent in a supervised manner as was done in Rebstock et al. for the game of \emph{Skat} \cite{rebstock2019learning}.
Backhus et al. examined the suitability of RL for the trick-taking card game \emph{Wizard} \cite{backhus2013application}.
In bidding, two decision-makers were implemented: one rule-based agent and one agent based on an FNN.
The FNN was trained in a supervised manner by approximating the tricks actually received at the end of each round.
In playing, the authors used a model-based RL approach originally proposed by Fujita et al. \cite{fujita2007model}.

Regarding the historic preprocessing, Obenaus \cite{obenausimplementing} analyzed the trick-taking game \emph{Doppelkopf}.
An LSTM was implemented to predict the next card going to be played using the history as input.

For improving strategies at decision-time, Monte-Carlo Tree Search (MCTS) is a promising path, but it cannot directly be applied to imperfect-information games.
Niklaus et al. \cite{niklaus2020challenging} instead evaluated two possible approaches called Determinized MCTS and Information Set MCTS\cite{Cowling2012ISMCTS} on \emph{Jass}.
Similar works have been conducted by Brown et al. \cite{brown2020combining} for \emph{Poker}, by Ishi et al. \cite{ishii2005reinforcement} for \emph{Hearts} and by Buro et al. \cite{buro2009improving} for \emph{Skat}.
All these works have in common that, before applying MCTS techniques, a state estimation must be conducted.
To that goal, Solinas et al. \cite{solinas2019improving} used both historic and general information as input to an FNN to predict the distribution of cards.
\section{Approach}

After clarifying general assumptions about the agent-environment interaction all three implemented models will be explained.





\subsection{Assumptions about the Agent-Environment Interaction} \label{assumption}

In Wizard, the final score of each player is determined by the sum of points received in all $15$ game rounds. It is therefore assumed that agents always try to attain as many points as they can in each round in order to win the overall game. This allows analyzing game rounds independently instead of monitoring the overall number of victories. 

Due to the particular properties of the reward function explained in the first section of this work, instead of the actual reward received at the end of each game round, the accuracy between bid and trick count is used to visualize the agents' performance.
This type of evaluation was chosen because it always ranges between 0\% and 100\% and therefore allows to compare performances across different game rounds.

Figure \ref{fig:structure_game_round} shows how each game round is modeled by two interconnected (partially observable) MDPs.
Bidding is a \emph{one-step MDP} in which the agent uses a fully-connected neural network (FNN) to map an input vector consisting of its own cards, its position in the player order, the trump suit, and the bids of the previous players to an action which is interpreted as that player's bid.
Playing is a \emph{multi-step MDP} containing as many cycles as there are cards to be played.
Apart from the information already used in bidding, the input vector now also contains the number of tricks already taken in previous tricks, the player's position, the suit to follow, and the highest card in the current trick.
The output of the playing network is interpreted as the card to be played in that specific trick.
As the player's choice of which card to play is restricted most of the time, the possible actions, varying in count throughout one round, have to be filtered using invalid action masking~\cite{Cheng2020masking}.
Elements of the input vectors are one-hot encoded.
The bidding and playing MDPs share the same reward signal which corresponds to the [0,1]-normalized points the agents receive at the end of the round.

\subsection{Deep Q-Networks}

Training is done using the DQN algorithm introduced by Mnih et al. \cite{mnih2015human}.
The underlying implementation\footnote{\url{https://github.com/jonas-schumacher/drl-pomdp}} 
is based on reference implementations from \cite{morales2020grokking} and \cite{lapan2020deep}.

\begin{figure}
\vspace{-0.15in}
	\centering
	\includegraphics[width=.35\textwidth]{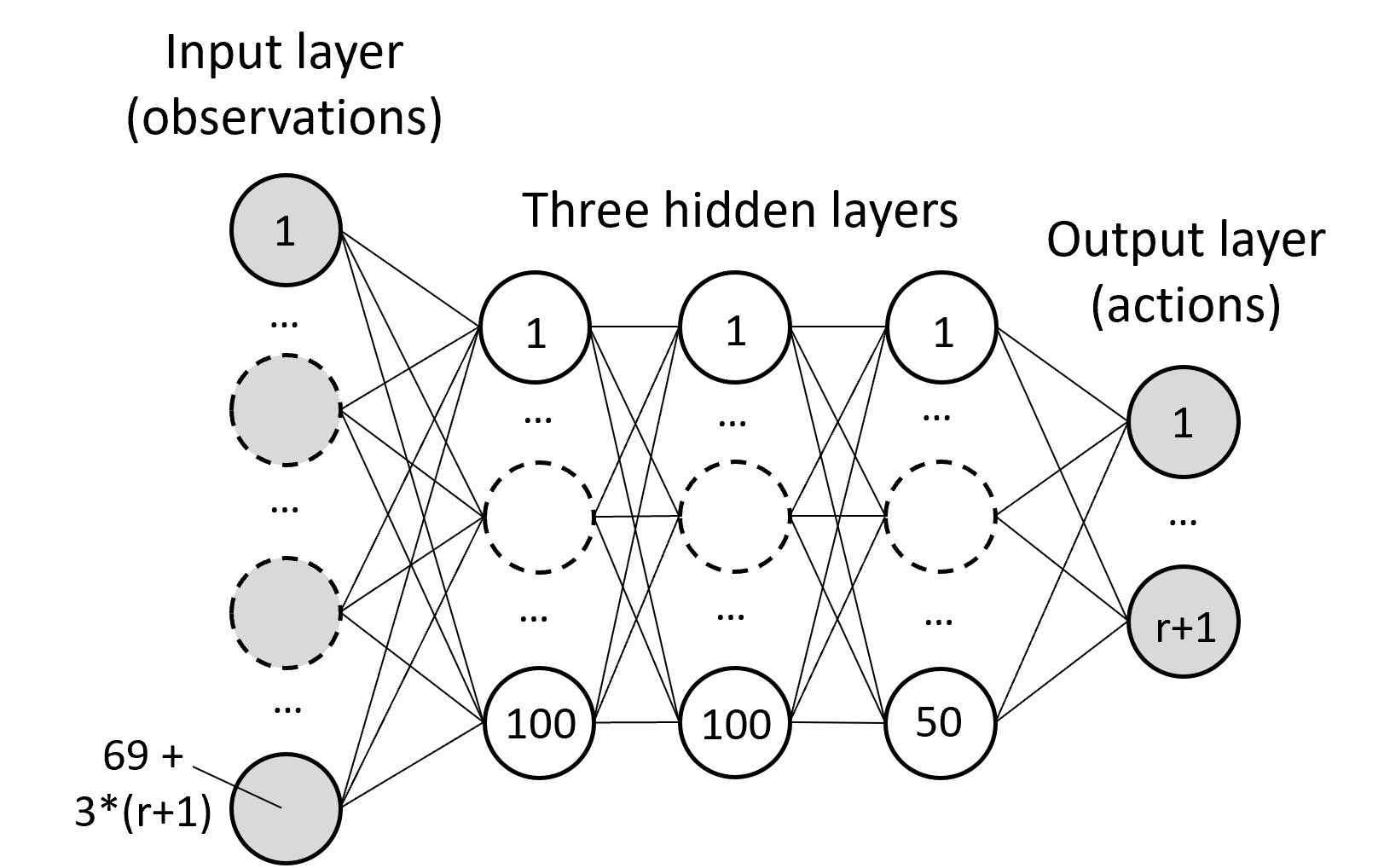}
	\caption{Structure of the fully-connected neural network in the bidding phase. For each of the $r\in \{1,2,...,15\}$ game rounds a separate network with a different input and output shape is trained.}
	\label{fig:fnn_bidding}
\end{figure}

\begin{figure}
\vspace{-0.15in}
	\centering
	\includegraphics[width=.4\textwidth]{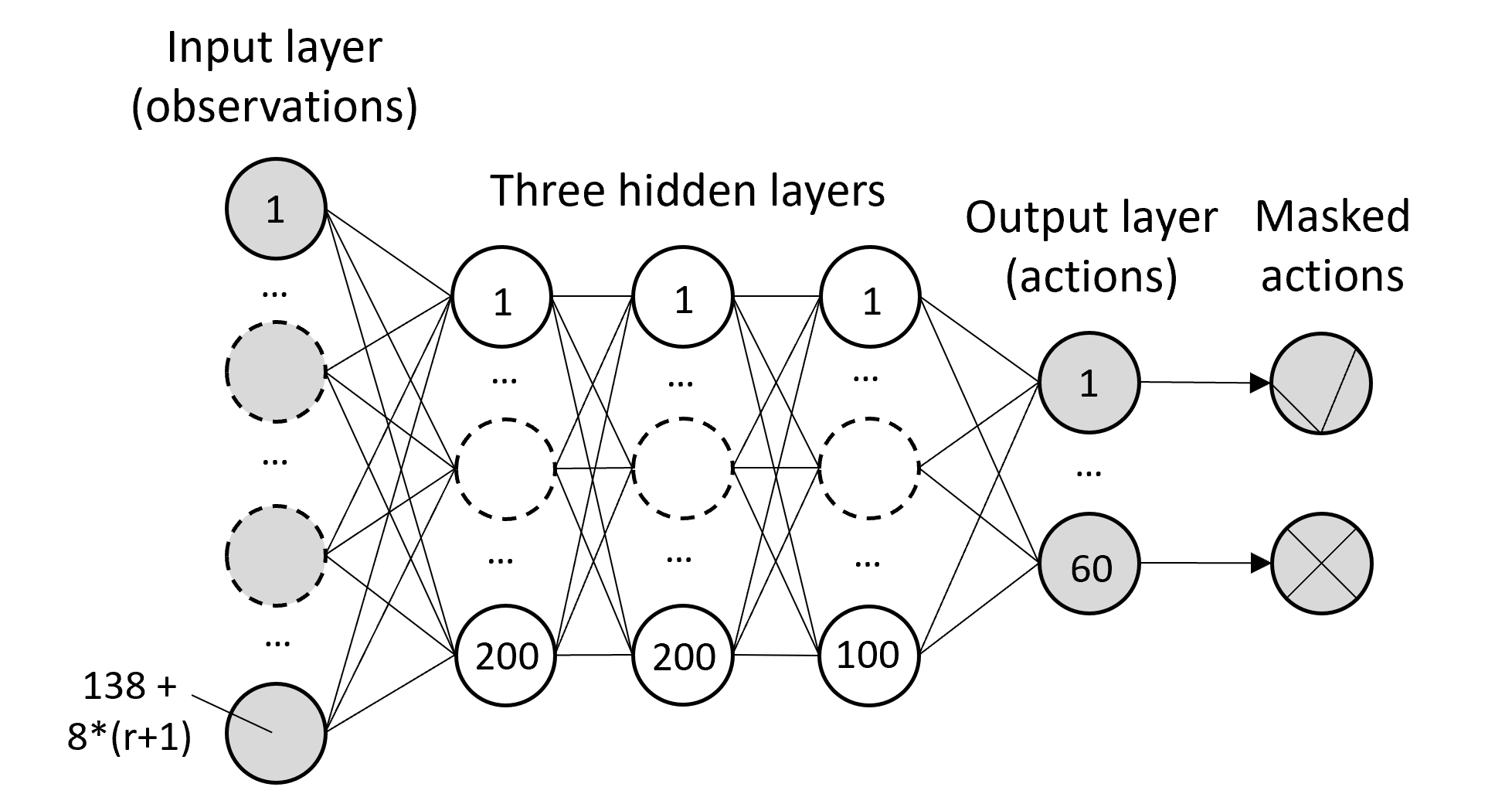}
	\caption{Structure of the fully-connected neural network in the playing phase. For each of the $r\in \{1,2,...,15\}$ game rounds a separate network with a different input shape is trained. The output layer is masked in order to remove inadmissible actions.}
	\label{fig:fnn_playing}
\vspace{-0.15in}
\end{figure}

Figures \ref{fig:fnn_bidding} and \ref{fig:fnn_playing} show the architecture of the neural networks that are used for bidding and playing respectively. For each of the 15 game rounds, two (one for bidding and one for playing) FNN with three hidden layers are trained.
The training is conducted in self-play.
While a game round proceeds, all four players use the same neural network for decision-making and the same replay buffer to save their individual experience tuples.
After each 10th game round the joint network is trained and after each 20th game round, 10\% of the target network weights are replaced by weights from the trained network.

\subsection{Historic Preprocessing}

The basic DQN algorithm presented in the previous subsection implicitly contains some part of the game's history encoded in the input vector.
In this subsection, the whole history of played cards will be processed by a long short-term memory (LSTM) network \cite{Hochreiter1997}.
The hidden state from that LSTM network will then be used as additional input to the DQN algorithm.

\begin{figure}
\vspace{-0.15in}
	\centering
	\includegraphics[width=.45\textwidth]{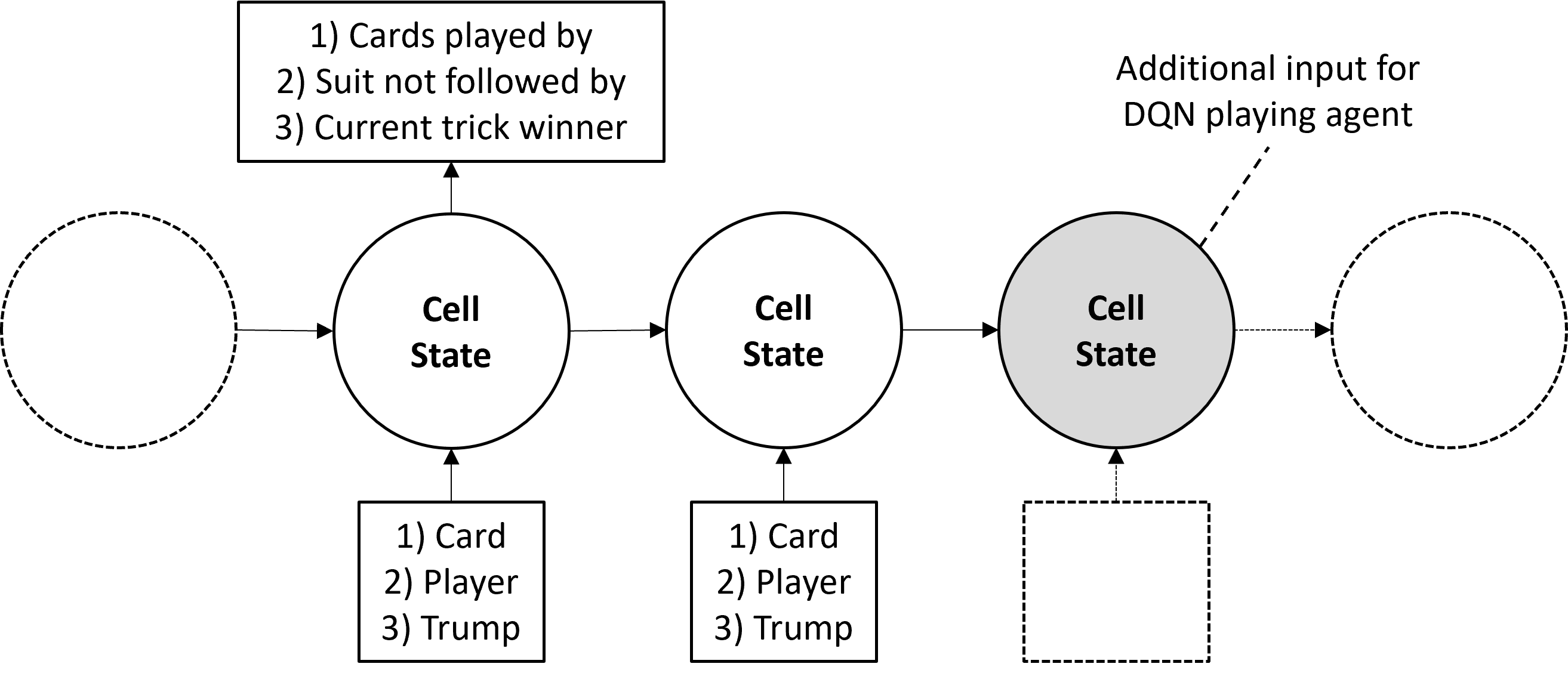}
	\caption{Unfolded LSTM cell representing the historic preprocessing. The input sequence of played cards is mapped to an output sequence of specific tasks. The hidden cell state is later used as additional input for the DQN agent.}
	\label{fig:lstm}
\vspace{-0.15in}
\end{figure}

Figure \ref{fig:lstm} shows the learning task of the LSTM which corresponds to a sequence-to-sequence model.
The input sequence consists of the card that has been played, the player who played it and the current trump suit.
The output sequence also consists of three elements: the information about which player played which cards so far, which suit cannot be followed by which player and which player played the highest card in the current trick.

\begin{figure}
	\centering
	\includegraphics[width=.4\textwidth]{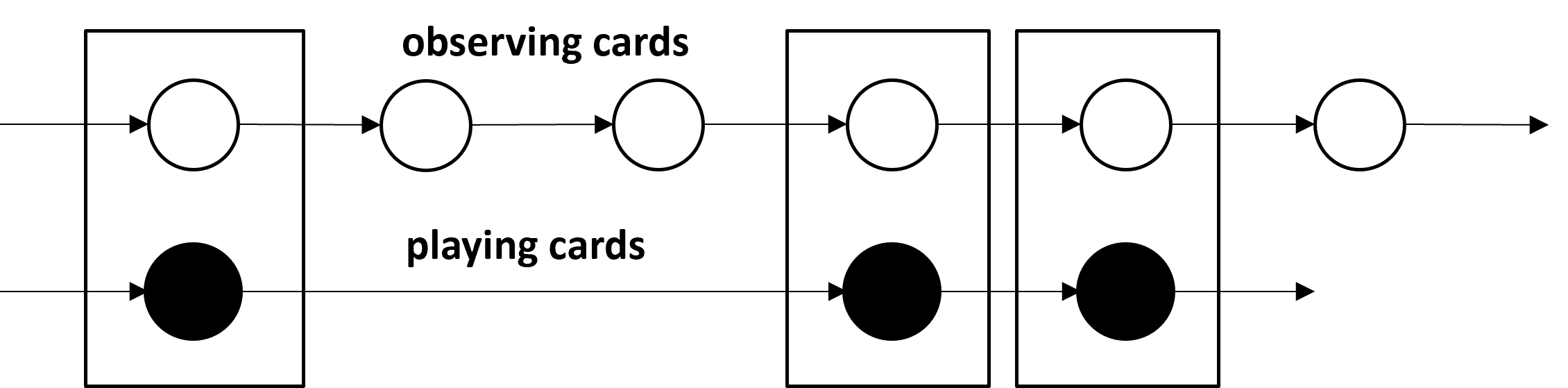}
	\caption{Different time scales in observing cards (used by the LSTM agent in historic preprocessing) compared to playing cards (used by the DQN agent in decision-making).}
	\label{fig:lstm_scale}
\vspace{-0.15in}
\end{figure}

By processing the whole history of played cards, the agent acts on the white time scale shown in figure \ref{fig:lstm_scale}.
This means the agent observes played cards even if it is not asked for action.
This contrasts with the basic DQN agent which acts on the black time scale in figure \ref{fig:lstm_scale}. 

\subsection{Model-based Tree Search}

For the last approach covered in this paper, the agent performs a forward-directed tree search at decision-time to improve its choices in the playing phase.
The approach has been inspired by \cite{niklaus2020challenging} and \cite{ishii2005reinforcement}.
To establish such a search the agent needs a model of the environment, which means it must know or approximate how its actions affect both the state transition and the reward function of the environment.

\begin{figure}
	\centering
	\includegraphics[width=.35\textwidth]{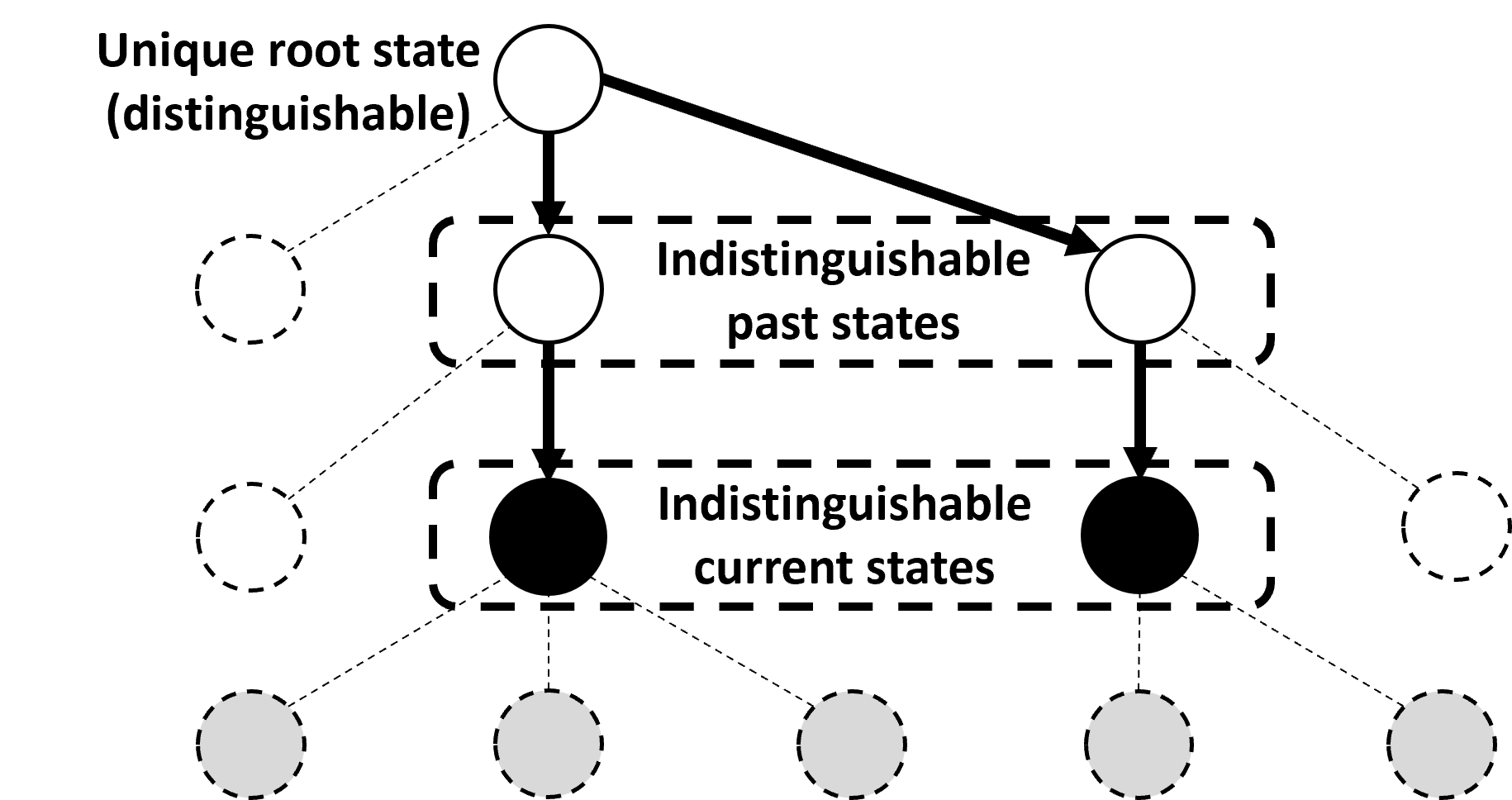}
	\caption{A generic game tree with two possibly realized paths shown in bold black. The grey nodes depict possible future game states.}
	\label{fig:game_tree_generic}
\vspace{-0.2in}
\end{figure}

However, in trick-taking games, the agent has uncertainty about two components of the environment. First, the agent does not know the true state of the environment (i.e., its position in the game tree) due to the partial observability of the environment. This is visualized in figure \ref{fig:game_tree_generic}. The agent cannot conduct a tree search to explore the grey nodes if it does not know in which of the black nodes it is actually located. Second, as the agent's opponents are also part of the environment, it does not know with certainty how they will behave during the game. Therefore, the agent needs to pretend to have full information about both the state and the strategy of the other players.

\begin{table}
\caption{The state of the environment can be formulated by mapping each card to a specific location.}
\begin{center}
\begin{tabular}{|c|c|c|c|c|c|c|c|c|}
\hline
 & In the deck & \multicolumn{3}{|c|}{In the hand of} & \multicolumn{2}{|c|}{Played by} & $\Sigma$
\\ \hline
 &  & P1 & P2 & $\cdots$ & P1 & $\cdots$ & 
\\ \hline
Card 1 & 0 & 1 & 0 & $\cdots$ & 0 & $\cdots$ & 1
\\ \hline
Card 2 & 1 & 0 & 0 & $\cdots$ & 0 & $\cdots$ & 1
\\ \hline
$\cdots$ & $\cdots$ & $\cdots$ & $\cdots$ & $\cdots$ & $\cdots$ & $\cdots$ & $\cdots$
\\ \hline
Card 60 & 0 & 0 & 1 & $\cdots$ & 0 & $\cdots$ & 1
\\ \hline
\end{tabular}
\label{table:state_model}
\end{center}
\vspace{-0.20in}
\end{table}

Regarding the first challenge, a model of the environment can be described by the location of each card as shown in table \ref{table:state_model}. Once the agent knows one specific state, it can simulate subsequent states of the game by observing the other players' actions. In this work, three possible ways of creating a model have been implemented:
\begin{enumerate}
\item Using the ground truth from the environment
\item Sampling from a uniform distribution of possible states
\item Sampling based on the output of a neural network
\end{enumerate}

Regarding the second challenge, it was assumed that the other players follow the same strategy as the optimized player. In the case of the self-play algorithm implemented in this work, this assumption even corresponds to the truth.

Once the agent has sampled a state of the environment and has an assumption about how the other players act, it can perform a tree search. To this end, whenever the agent is called upon action in the playing phase, it simulates the environment and finishes the current game round. If another player is asked to choose an action during that simulation, the agent uses its own DQN network to simulate that decision. It performs one simulation for each possible action and chooses the action that results in the highest reward.
\section{Experimental Analysis}

The experiments in this section are divided into three subsections. 
First, the results of the basic DQN agent are presented.
Second, the historic preprocessing using an LSTM is shown.
And third, the optimization at decision-time based on a model of the environment is illustrated.



\subsection{Deep Q-Networks}


The hyperparameters that are used in DQN are mostly identical for bidding and playing.
They are inspired by the aforementioned reference implementations and are further tuned using trial-and-error.
Notably, a much larger replay buffer and batch size are vital.
\begin{itemize}
\item Gamma ($\gamma$): $1.0$
\item Exploration strategy: exponential $\varepsilon$-decay from $1.0$ to $0.01$ in $90$\% of training time
\item Replay buffer size: $300,000$ (bidding) / $600,000$ (play.)
\item Batch size: $1024$
\item Activation function in hidden layers: ReLU
\item Learning rate: $0.0005$
\item Optimizer: Adam
\end{itemize}


\begin{figure}
    \centering
	\includegraphics[width=0.325\textwidth]{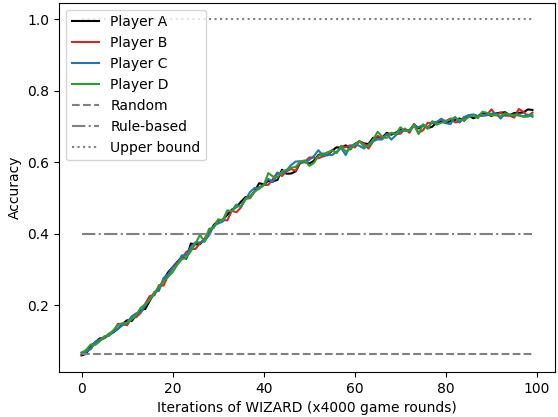}
    \caption{Accuracy of identical DQN agents during self-play training of round 15 (last round). Performance differences across players are caused by the stochasticity of the environment.}
	\label{fig:training_process}
\vspace{-0.20in}
\end{figure}

As one example, figure \ref{fig:training_process} shows the accuracy during the training of the last round (round 15) of the DQN agents. Each data point shows the average accuracy of 4000 game rounds. Agents perform almost identically because bidding positions were chosen randomly and all agents use the the same neural networks for decision-making. 

\begin{figure}
    \centering
	\includegraphics[width=0.356\textwidth]{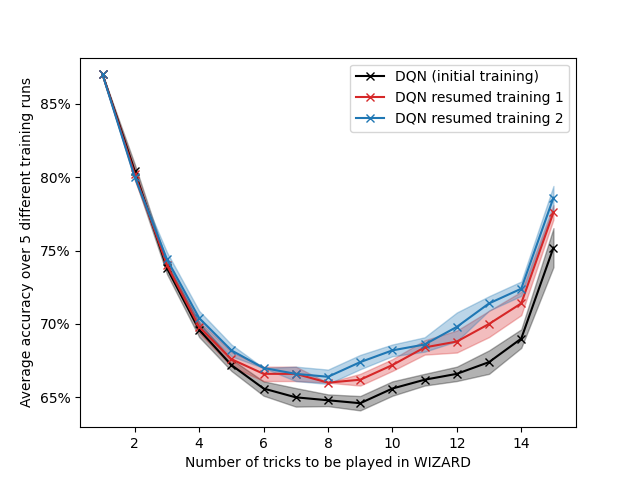}
    \caption{Mean and standard deviation of 5 different runs of the final accuracies for three different versions of the DQN algorithm.}
	\label{fig:seeds_improvement}
\vspace{-0.20in}
\end{figure}

Figure \ref{fig:seeds_improvement} shows the final accuracy of the DQN agent for all 15 game rounds. Each colored curve shows the mean and standard deviation of 5 optimization runs using different seeds. The performance of the DQN agent was measured in evaluation mode, which means an already trained agent with a purely greedy strategy  ($\varepsilon=0$) was used. Each data point is the average of 10,000 game rounds per bidding position which corresponds to 40,000 game rounds in total. The black chart is the initial DQN run and the red and blue charts are retrained improvements using checkpoints of previous DQN versions. For the retrained versions, training started with a lower exploration value of $\varepsilon=0.3$.
What is salient here is the pronounced U-shape of the final accuracies. It seems that the agent struggles most in intermediate rounds, where the final accuracy is around 65-67\%. In the first round, it is constantly on a level of 87\% and in the last round, it reaches levels of almost 80\%.


\begin{figure}
    \centering
	\includegraphics[width=0.356\textwidth]{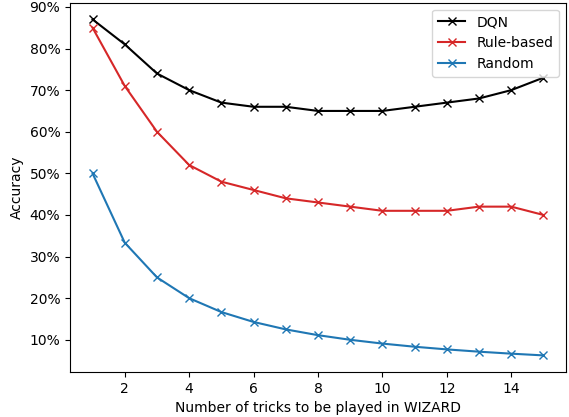}
    \caption{Accuracy of different agent types across game rounds.}
	\label{fig:wizard_accuracy}
\vspace{-0.10in}
\end{figure}

Figure \ref{fig:wizard_accuracy} compares the final accuracies of one of the initial DQN runs presented in figure \ref{fig:seeds_improvement} to that of a rule-based agent and a random agent each playing against three opponents of the same type. The random agent in blue always chooses actions at random and its performance strictly decreases from $\frac{1}{2}=50\%$ to $\frac{1}{16}\approx 6\%$. The rule-based agent uses hard-coded rules for decision-making which can be found in the contributed source code. In bidding, it evaluates its hand by calculating the winning probability of each card. In playing, it chooses an admissible card based on the difference of its bid and trick count and its chances to win the current trick. Notice that figure \ref{fig:wizard_accuracy} and all following figures represent the evaluation of one training run, i.e. using one seed only. This is motivated by figure \ref{fig:seeds_improvement} which conveyed the fact that results across multiple training runs show little variance.

\begin{figure}
    \centering
	\includegraphics[width=0.356\textwidth]{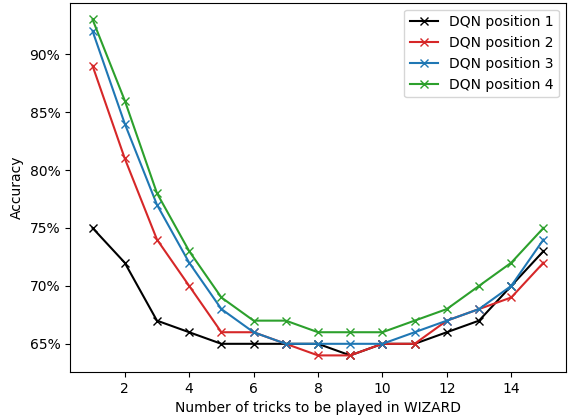}
    \caption{Influence of the player's bidding position on the final accuracy.}
	\label{fig:wizard_position}
\vspace{-0.2in}
\end{figure}

The black curve from figure \ref{fig:wizard_accuracy} represents an average over all possible player positions. It can further be divided into accuracies based on the bidding positions which are shown in figure \ref{fig:wizard_position}. The accuracy tends to increase with the player's position and this effect is especially strong in the first rounds of the game. 

\begin{figure}
\vspace{-0.20in}
    \centering
	\includegraphics[width=0.356\textwidth]{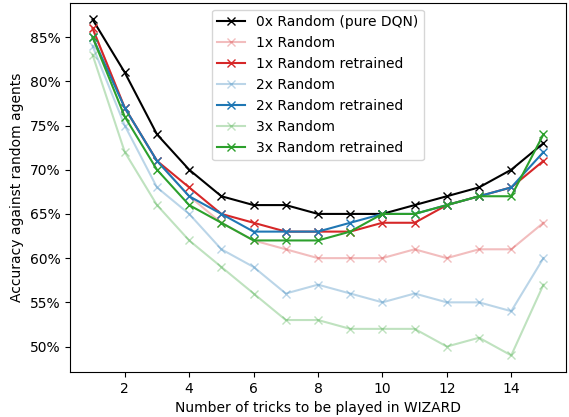}
    \caption{Performance of DQN agents against varying numbers of random opponents. Accuracies shown in light colors are based on the self-play DQN agent also shown in figure \ref{fig:wizard_accuracy}. For the solid curves, DQN agents were retrained against random opponents.}
	\label{fig:wizard_random}
\vspace{-0.10in}
\end{figure}

Figure \ref{fig:wizard_random} shows the performance of DQN agents against different numbers of random agents. The pale curves correspond to a direct evaluation of the trained DQN agents and decline with the number of random agents participating in the game. The solid curves are accuracies measured after an additional DQN training against random agents using the same hyperparameters as in self-play training. With this additional training, the performance is only slightly worse than that of the pure self-play DQN shown in black. Notice that this has no effect on the accuracy of the random agents which is constantly $\frac{1}{r+1}$ where $r$ is the round to be played.

\begin{figure}
    \centering
	\includegraphics[width=0.356\textwidth]{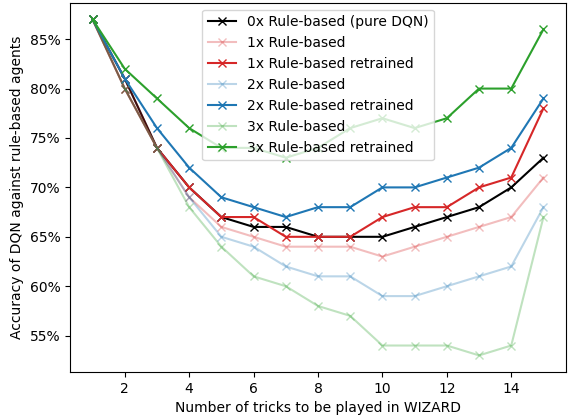}
    \caption{Performance of trained DQN agents against different numbers of rule-based opponents. Similar to figure \ref{fig:wizard_random}, light curves show accuracies of the self-play DQN agent whereas solid curves are retrained DQN versions.}
	\label{fig:wizard_rule_dqn}
\vspace{-0.20in}
\end{figure}

Figure \ref{fig:wizard_rule_dqn} shows similar charts as figure \ref{fig:wizard_random} for the evaluation against rule-based agents. Again, in the case of direct evaluation, the performance of the DQN agent drops if more rule-based agents participate in the game. However, even in the case of 3 rule-based opponents, the DQN agent achieves accuracies of more than 55\% without ever having played against rule-based agents before. This shows that the self-play training was able to generalize toward a wider range of possible opponents. If DQN agents are allowed to retrain themselves, their performance even surpasses that of the self-play DQN agent. This effect is strongest if more rule-based agents are involved. Surprisingly, playing against DQN agents has a positive effect on the accuracy of rule-based agents. Their average accuracy per game round increases by 2 (in the case of one DQN agent) to 6 (in the case of three DQN agents) percentage points compared to their baseline performance.

\subsection{Historic Preprocessing}

The LSTM training consists of two parts. First, the LSTM agent is trained in a supervised manner to learn a representation of the explicit history provided as input sequence. Second, in the playing phase, the internal state of the agent shown in figure \ref{fig:lstm} is concatenated to the input of the basic DQN agent.

During training, the following hyperparameters are used:
\begin{itemize}
\item Replay buffer size: $10,000$
\item Batch size: $64$
\item Size of hidden cell: 50 / 100 / 150
\item Learning rate: $0.005$
\end{itemize}

\begin{figure}
	\centering
	\includegraphics[width=.356\textwidth]{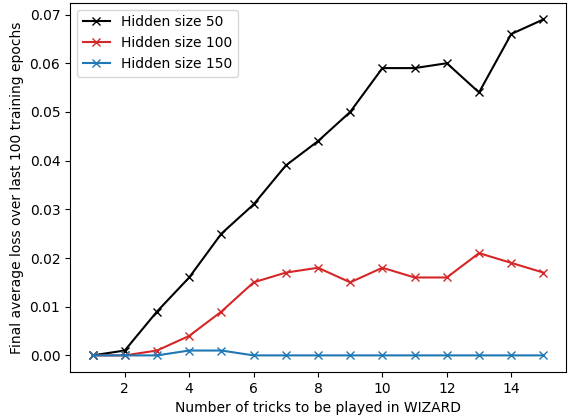}
	\caption{Final loss (averaged over 100 training epochs) of training the LSTM network across different game rounds.}
	\label{fig:history_loss}
\vspace{-0.10in}
\end{figure}

Figure \ref{fig:history_loss} shows the final loss of the LSTM training averaged over 100 training iterations as a function of all 15 game rounds. The three graphs represent the size of the hidden cell of the LSTM agent. As can be seen, a hidden size of 150 is sufficient to perfectly map the input sequence to the output sequence. This shows that a representation of the game's history is indeed learnable by an LSTM network.

\begin{figure}
	\centering
	\includegraphics[width=.356\textwidth]{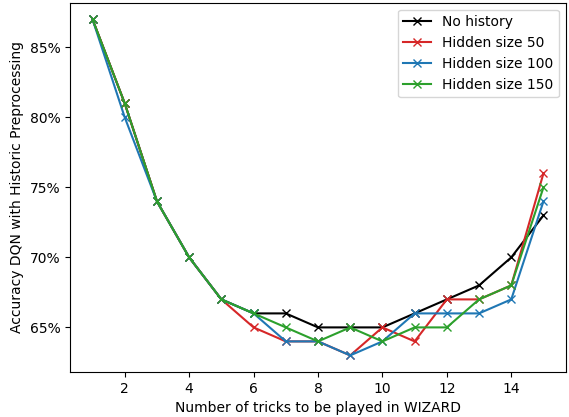}
	\caption{Final accuracy of DQN agents in self-play when using additional historic input from LSTM preprocessing.}
	\label{fig:dqn_hist}
\vspace{-0.15in}
\end{figure}

Figure \ref{fig:dqn_hist} shows the final accuracy of the DQN agent with additional historic input in comparison to the basic DQN agent without historic preprocessing. Except for the last round, accuracies including a historic representation do not exceed the basic DQN accuracy. On the contrary, including the history seems to deteriorate the agent's performance. In addition to that, the size of the hidden cell does not have a strong impact on the performance either.

\subsection{Model-based Tree Search}

The model-based approach comprises two parts: a state of the environment is sampled and then a tree search is conducted. For the state estimation using the ground truth and the uniform sampling method no training is required. 

For the sampling based on the output of a neural network, a fully-connected neural network was trained in a supervised manner. As input vector, it received the same information as the DQN agent and as output it received the true state of the environment. These hyperparameters are used for training:
\begin{itemize}
\item Replay buffer size: $600,000$
\item Batch size: $1024$
\item Shape of hidden layers: [200, 200, 300]
\item Learning rate: $0.001$
\end{itemize}

\begin{figure}
	\centering
	\includegraphics[width=.356\textwidth]{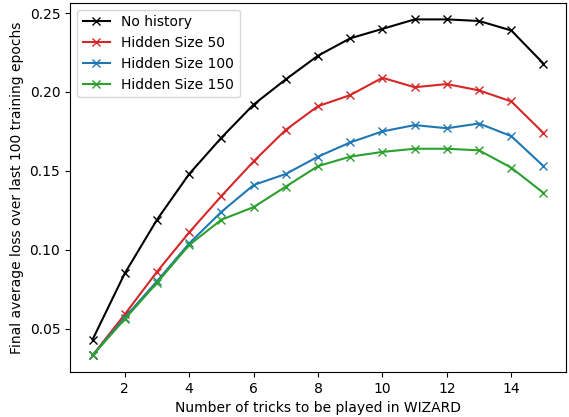}
	\caption{Final loss (averaged over 100 training epochs) of training the state estimator across different game rounds.}
	\label{fig:model_loss}
\vspace{-0.15in}
\end{figure}

Figure \ref{fig:model_loss} shows the final average loss across different game rounds. The black curve represents the loss which corresponds to the input of the basic DQN agent without historic preprocessing. The other curves represent runs with an extended input using information from the LSTM cell state presented in the previous subsection. Two observations can be made. First, the model of the environment gets more accurate the more historic information is available. And second, finding an accurate model seems to be most difficult for intermediate rounds.

\begin{figure}
	\centering
	\includegraphics[width=.356\textwidth]{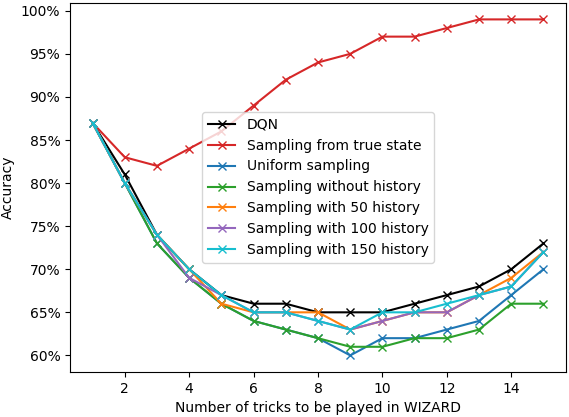}
	\caption{Final accuracy of a DQN agent performing a tree search in the playing phase.}
	\label{fig:model_accuracy}
\vspace{-0.20in}
\end{figure}

Figure \ref{fig:model_accuracy} shows the final accuracies when making decisions based on the results from the tree search. The first thing to notice is that using uniform sampling or sampling without historic information yields the worst results. But even sampling from the outputs of a neural network with access to historic information performs worse than the original DQN algorithm (shown in black). Therefore, no sampling method using imperfect-information was able to surpass the basic DQN performance. Using the true state (shown in red) is of course no realistic scenario because it pretends that the agent has perfect information. However, the gap between the red curve and the other curves can be interpreted as the value of that missing information. It means that if the agent had perfect information in the playing phase, it could reach accuracies of up to 99\% in later rounds of the game. What is equally interesting is the gap between the red curve and the upper bound of 100\%. This gap represents the imperfect bidding decisions, which apparently cannot be corrected by the playing component of the agent.
\section{Discussion}
Charlesworth et al. point out that ``approaches such as Deep Q-Networks struggle because multi-agent environments are inherently non-stationary (due to the fact that the other agents are themselves improving with time) which prevents the straightforward use of experience replay that is necessary to stabilize the algorithm"~\cite{charlesworth2018application}.
However, the results presented in this work do not confirm that claim because the self-play DQN algorithm was indeed able to perform well on the DRL problem at hand.

The performance of the DQN agent and the other two approaches as well as potential consecutive work are subsequently discussed.


\subsection{Emerged Behaviors and the Value of Information}

When comparing across different game rounds, the performance anomaly (i.e., the u-shaped curve) in figure \ref{fig:seeds_improvement} is especially salient.
The initial decrease in performance is not surprising because both random and rule-based baselines in figure \ref{fig:wizard_accuracy} have similar shapes.
For the later increase, there could be several explanations.
The first is that there is more knowledge in later game rounds. This is also supported by the model-based approach in which the agent could more easily learn a model of rounds 14 and 15 compared to rounds 12 and 13.
More information means a better understanding of the current situation and therefore a better performance.
The second explanation could be that the behavior of the other agents is more predictable in later rounds because most lead suits have to be followed.
This would also explain the especially good performance in the last round, in which no trump suit exists.
And third, it might be the case that agents have more control in the playing phases of later rounds because they involve more tricks and therefore more decisions.
This last explanation is supported by figure \ref{fig:model_accuracy} of the model-based approach, where it has been shown that an omniscient player gets increasingly better the more decisions it can take in the playing phase. 

What has been shown in figure \ref{fig:wizard_position} is the additional value of having a later position in bidding.
Especially in game rounds with fewer tricks to be played, players performed significantly better if they knew the other players' bids.
Calculating the difference between the performance of identical players allowed to quantify the value of information in imperfect-information games.

Another interesting result is the evaluation against rule-based agents in figure \ref{fig:wizard_rule_dqn}. It showed that DQN agents struggled when playing against rule-based agents they had never faced before.
But given another training iteration, their performance even surpassed their basic performance with no rule-based agents involved.
This could be explained by the fact that rule-based agents are more predictable than DQN agents and the environment becomes more stationary if less self-improving DQN agents are involved.
The same analysis revealed that, regardless of whether DQN agents were retrained or not, rule-based agents performed better the more DQN agents were involved. 
This shows that, at least under the given assumption of maximizing points instead of ranks, the Wizard environment is not a strictly competitive, zero-sum game like Chess or Go.
If the total number of bids equals the number of tricks to be played in a specific game round, agents might be incentivized to help other agents reach their initial bid because everyone profits from attaining a higher final accuracy.

\subsection{The Effect of Maximizing Points on the Winning Probability}

In the approach section it was argued that maximizing the points in each individual round would also increase the probability of winning the overall game. To support this claim, the fully trained DQN agent was evaluated in $10,000$ full games of $15$ rounds.

\begin{figure}
    \centering
	\includegraphics[width=0.325\textwidth]{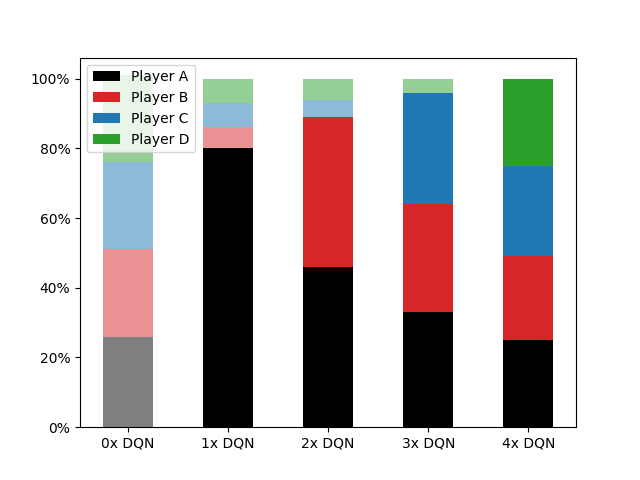}
    \caption{Winning share of DQN agents (solid bars) and rule-based agents (light bars) in $10,000$ full games of 15 rounds.}
	\label{fig:winning_share}
\vspace{-0.15in}
\end{figure}

Figure \ref{fig:winning_share} shows the winning share for a varying number of DQN agents playing against rule-based agents (after having been retrained against them).

Unsurprisingly, the winning share of 4 rule-based agents (left stacked bar) and 4 DQN agents (right stacked bar) in self-play is approximately $\frac{1}{4}=25\%$). When one DQN agent (Player A) participates in the game, it wins in 80\% of all games while its opponents reach $6 + 7 + 7 = 20\%$. In the case of two DQN agents (Player A \& Player B) their share amounts to  $46 + 43 = 89\%$ against $5 + 6 = 11\%$ for their rule-based counterparts.

This evaluation conveys two important findings. First, maximizing points in individual rounds proved to be a good objective function for also winning the overall game. Second, in mixed-player games all agents perform worse the closer they are positioned behind other DQN agents.

When evaluated against random agents, DQN agents' winning share amounts to more than 99.9\% across all player constellations leaving random agents chanceless.

\subsection{Importance of Historic Input}

Regarding the historic preprocessing, the LSTM network proved to be a good architecture to process the game's history on a card-wise basis.
The model was able to learn three diverse tasks showing its ability to understand the trick structure and value of cards without being explicitly programmed.
However, using the internal representation of the LSTM cell state as additional input to the DQN agent did not improve its overall performance.
There could be several reasons for this.
It could be the case that the DQN agent already had enough historic information implicitly contained in its observational input.
Another case could be that it was more difficult for the DQN agent to extract the relevant information from a now more extensive input vector.
Increasing the network size and learning time might improve the results.
And finally, the tasks to be solved by the LSTM might not have been fully representative of the game's history.
In potential future work, the LSTM network could be trained with different and more diverse tasks.
Another approach would be to directly combine the DQN agent with a recurrent neural network which is referred to as Deep Recurrent Q-Networks in literature \cite{HausknechtS15DRQN}.

\subsection{Improvement at Decision-time using Tree Search}

The model-based tree search approach could also not improve the DQN results but revealed interesting properties of the underlying game.
Figure \ref{fig:model_loss} showed that the amount of information available to the agent first decreased and then increased again for later game rounds.
This could be due to the fact that in the first game rounds, the agent can be certain that most cards remain undealt in the deck.
The more cards are distributed among other players, the more difficult it gets for the player to know where each card is located.
This effect is reversed in later game rounds when the agent can be increasingly sure that no cards remain in the deck and can also remember which cards have already been played.
Comparing the DQN agent to an omniscient player in figure \ref{fig:model_accuracy} further visualized the lack of the DQN agent to take optimal decisions in the playing phase.
What would be worth investigating in the future is a way of combining the DQN decision-maker with the model-based decision-maker.
If little information is available, e.g., in the bidding phase and during the first tricks, the DQN agent could decide.
In later tricks, when more information is available, the tree-search component could take over.
\section{Conclusion}

Due to the many challenges, the trick-taking game Wizard analyzed in this paper is an intriguing testbed for studying DRL algorithms.
The DQN agent showed a consistent improvement during the self-play training of its bidding and playing component. Notably, the trained DQN agent is robust and able to generalize to an unseen rule-based agent, while dealing with the highly non-stationary multi-agent setting.
If trained against that rule-based agent, it was able to increase its own and its opponents' performance simultaneously.
This conveyed the insight that the analyzed environment is not strictly competitive.
A deeper analysis of the player's bidding position revealed the value of knowing the bids of the previous players.
Independently from each other, both the DQN and the model-based training showed a performance anomaly across different game rounds.
After an initial drop, the performance increased again, which might be linked to the amount of information available to the agents.
The surprising findings on the historic preprocessing and model-based approach leave potential for future work.



\bibliographystyle{./bibliography/IEEEtran}
\bibliography{./bibliography/IEEEabrv.bib,./bibliography/bibliography.bib}

\begin{thebibliography}{10}
\providecommand{\url}[1]{#1}
\csname url@samestyle\endcsname
\providecommand{\newblock}{\relax}
\providecommand{\bibinfo}[2]{#2}
\providecommand{\BIBentrySTDinterwordspacing}{\spaceskip=0pt\relax}
\providecommand{\BIBentryALTinterwordstretchfactor}{4}
\providecommand{\BIBentryALTinterwordspacing}{\spaceskip=\fontdimen2\font plus
\BIBentryALTinterwordstretchfactor\fontdimen3\font minus
  \fontdimen4\font\relax}
\providecommand{\BIBforeignlanguage}[2]{{%
\expandafter\ifx\csname l@#1\endcsname\relax
\typeout{** WARNING: IEEEtran.bst: No hyphenation pattern has been}%
\typeout{** loaded for the language `#1'. Using the pattern for}%
\typeout{** the default language instead.}%
\else
\language=\csname l@#1\endcsname
\fi
#2}}
\providecommand{\BIBdecl}{\relax}
\BIBdecl

\bibitem{mnih2015human}
V.~Mnih, K.~Kavukcuoglu, D.~Silver, A.~A. Rusu, J.~Veness, M.~G. Bellemare,
  A.~Graves, M.~Riedmiller, A.~K. Fidjeland, G.~Ostrovski \emph{et~al.},
  ``Human-level control through deep reinforcement learning,'' \emph{Nature},
  vol. 518, no. 7540, pp. 529--533, 2015.

\bibitem{silver2017mastering}
D.~Silver, J.~Schrittwieser, K.~Simonyan, I.~Antonoglou, A.~Huang, A.~Guez,
  T.~Hubert, L.~Baker, M.~Lai, A.~Bolton \emph{et~al.}, ``Mastering the game of
  go without human knowledge,'' \emph{Nature}, vol. 550, no. 7676, pp.
  354--359, 2017.

\bibitem{sutton2018reinforcement}
R.~S. Sutton and A.~G. Barto, \emph{Reinforcement learning: An introduction},
  2nd~ed.\hskip 1em plus 0.5em minus 0.4em\relax MIT press, 2018.

\bibitem{morales2020grokking}
M.~Morales, \emph{Grokking deep reinforcement learning}.\hskip 1em plus 0.5em
  minus 0.4em\relax Simon and Schuster, 2020.

\bibitem{yeh2018automatic}
C.-K. Yeh, C.-Y. Hsieh, and H.-T. Lin, ``Automatic bridge bidding using deep
  reinforcement learning,'' \emph{IEEE Transactions on Games}, vol.~10, no.~4,
  pp. 365--377, 2018.

\bibitem{baykal2019reinforcement}
O.~Baykal and F.~N. Alpaslan, ``Reinforcement learning in card game
  environments using monte carlo methods and artificial neural networks,'' in
  \emph{2019 4th International Conference on Computer Science and Engineering
  (UBMK)}.\hskip 1em plus 0.5em minus 0.4em\relax IEEE, 2019, pp. 1--6.

\bibitem{rebstock2019learning}
D.~Rebstock, C.~Solinas, and M.~Buro, ``Learning policies from human data for
  skat,'' in \emph{2019 IEEE Conference on Games (CoG)}.\hskip 1em plus 0.5em
  minus 0.4em\relax IEEE, 2019, pp. 1--8.

\bibitem{backhus2013application}
J.~C. Backhus, H.~Nonaka, T.~Yoshikawa, and M.~Sugimoto, ``Application of
  reinforcement learning to the card game wizard,'' in \emph{2013 IEEE 2nd
  Global Conference on Consumer Electronics (GCCE)}.\hskip 1em plus 0.5em minus
  0.4em\relax IEEE, 2013, pp. 329--333.

\bibitem{fujita2007model}
H.~Fujita and S.~Ishii, ``Model-based reinforcement learning for partially
  observable games with sampling-based state estimation,'' \emph{Neural
  computation}, vol.~19, no.~11, pp. 3051--3087, 2007.

\bibitem{obenausimplementing}
J.~Obenaus, ``Implementing a {D}oppelkopf card game playing ai using neural
  networks,'' Bachelor thesis:
  \url{https://www.mi.fu-berlin.de/inf/groups/ag-ki/Theses/Completed-theses/Bachelor-theses/2017/Obenaus/BA-Obenaus.pdf},
  2017, [Online; accessed 28-February-2022].

\bibitem{niklaus2020challenging}
J.~Niklaus, M.~Alberti, R.~Ingold, M.~Stolze, and T.~Koller, ``Challenging
  human supremacy: Evaluating monte carlo tree search and deep learning for the
  trick taking card game jass,'' in \emph{International Conference on
  Artificial Intelligence and Soft Computing}.\hskip 1em plus 0.5em minus
  0.4em\relax Springer, 2020, pp. 505--517.

\bibitem{Cowling2012ISMCTS}
P.~I. Cowling, E.~J. Powley, and D.~Whitehouse, ``Information set monte carlo
  tree search,'' \emph{{IEEE} Trans. Comput. Intell. {AI} Games}, vol.~4,
  no.~2, pp. 120--143, 2012.

\bibitem{brown2020combining}
N.~Brown, A.~Bakhtin, A.~Lerer, and Q.~Gong, ``Combining deep reinforcement
  learning and search for imperfect-information games,'' in \emph{Conference on
  Neural Information Processing Systems (NeurIPS)}, 2020.

\bibitem{ishii2005reinforcement}
S.~Ishii, H.~Fujita, M.~Mitsutake, T.~Yamazaki, J.~Matsuda, and Y.~Matsuno, ``A
  reinforcement learning scheme for a partially-observable multi-agent game,''
  \emph{Machine Learning}, vol.~59, no.~1, pp. 31--54, 2005.

\bibitem{buro2009improving}
M.~Buro, J.~R. Long, T.~Furtak, and N.~Sturtevant, ``Improving state
  evaluation, inference, and search in trick-based card games,'' in
  \emph{Twenty-First International Joint Conference on Artificial
  Intelligence}, 2009.

\bibitem{solinas2019improving}
C.~Solinas, D.~Rebstock, and M.~Buro, ``Improving search with supervised
  learning in trick-based card games,'' in \emph{Proceedings of the AAAI
  Conference on Artificial Intelligence}, vol.~33, 2019, pp. 1158--1165.

\bibitem{Cheng2020masking}
C.~Tang, C.~Liu, W.~Chen, and S.~D. You, ``Implementing action mask in proximal
  policy optimization {(PPO)} algorithm,'' \emph{{ICT} Express}, vol.~6, no.~3,
  pp. 200--203, 2020.

\bibitem{lapan2020deep}
M.~Lapan, \emph{Deep reinforcement learning hands-on}.\hskip 1em plus 0.5em
  minus 0.4em\relax Packt publishing, 2020.

\bibitem{Hochreiter1997}
S.~Hochreiter and J.~Schmidhuber, ``{Long Short-Term Memory},'' \emph{Neural
  Computation}, vol.~9, no.~8, pp. 1735--1780, 11 1997.

\bibitem{charlesworth2018application}
H.~Charlesworth, ``Application of self-play deep reinforcement learning to "big
  2", a four-player game of imperfect information,'' in \emph{AAAI-19: Workshop
  on Reinforcement Learning in Games (RLG)}, 2019.

\bibitem{HausknechtS15DRQN}
M.~J. Hausknecht and P.~Stone, ``Deep recurrent q-learning for partially
  observable mdps,'' in \emph{2015 {AAAI} Fall Symposia, Arlington, Virginia,
  USA, November 12-14, 2015}.\hskip 1em plus 0.5em minus 0.4em\relax {AAAI}
  Press, 2015, pp. 29--37.

\end{thebibliography}

\end{document}